\documentclass[10pt,twocolumn,letterpaper]{article}

\usepackage{cvm}
\usepackage{times}
\usepackage{epsfig}
\usepackage{graphicx}
\usepackage{amsmath}
\usepackage{amssymb}
\usepackage{caption}

\graphicspath{{./imgs/}}
\DeclareGraphicsExtensions{.pdf, .jpg, .png}

\usepackage{threeparttable}
\usepackage{amsfonts} 
\usepackage{stmaryrd}
\usepackage{multirow}
\usepackage{booktabs}
\usepackage{bm}

\usepackage{array}
\newcommand{\PreserveBackslash}[1]{\let\temp=\\#1\let\\=\temp}
\newcolumntype{C}[1]{>{\PreserveBackslash\centering}p{#1}}
\newcolumntype{R}[1]{>{\PreserveBackslash\raggedleft}p{#1}}
\newcolumntype{L}[1]{>{\PreserveBackslash\raggedright}p{#1}}

\usepackage{algorithm}
\usepackage{algpseudocode}

\usepackage[pagebackref=true,breaklinks=true,letterpaper=true,colorlinks,bookmarks=false]{hyperref}

 \cvmfinalcopy 

\ifcvmfinal\fi
\begin{document}

\title{
    Cloud Sphere: A 3D Shape Representation via Progressive Deformation
    }


\author{Zongji Wang\\
Key Laboratory of Network Information System Technology (NIST), \\
Aerospace Information Research Institute, \\
Chinese Academy of Sciences,\\
Beijing 100190, China.\\
{\tt\small wangzongji@aircas.ac.cn}
\and
Yunfei Liu\\
State Key Laboratory of Virtual Reality Technology and Systems, \\
School of Computer Science and Engineering, \\
Beihang University, Beijing 100191, China. \\
{\small\url{lyunfei@buaa.edu.cn}}
\and
Feng Lu\\
State Key Laboratory of Virtual Reality Technology and Systems, \\
School of Computer Science and Engineering, \\
Beihang University, Beijing 100191, China. \\
Peng Cheng Laboratory, Shenzhen 518000, China. \\
{\tt\small lufeng@buaa.edu.cn}
}

\twocolumn[{
    \renewcommand\twocolumn[1][]{#1}
    \maketitle
    \begin{center}
        \centering
        \includegraphics[width=0.94\textwidth]{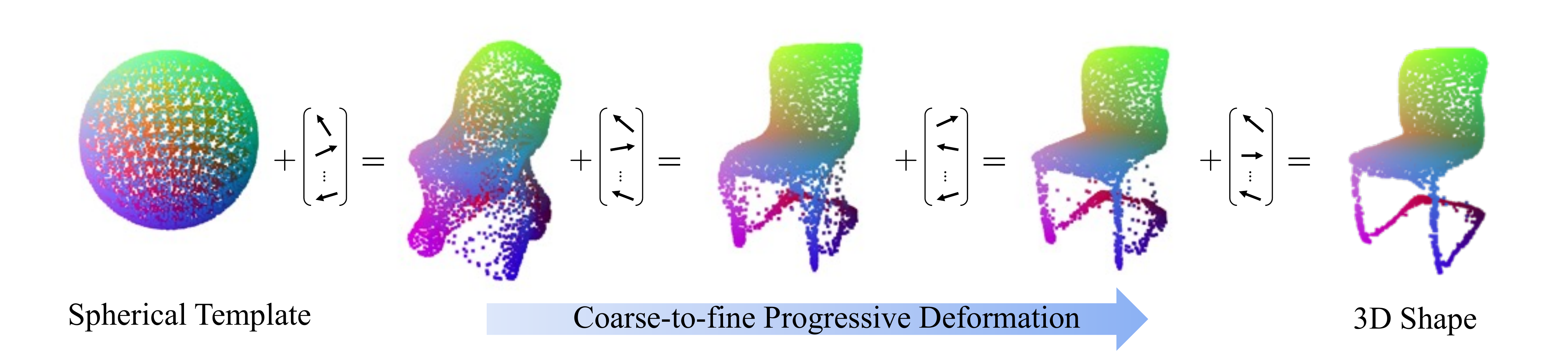}
        \captionof{figure}{The coarse-to-fine progressive deformation process for shape reconstruction from a spherical template. The distinctive information of the shape is progressively added throughout the shape formation process.}
        \label{teaser}
    \end{center} 
}]

\maketitle

\begin{abstract}
    In the area of 3D shape analysis, the geometric properties of a shape have long been studied. Instead of directly extracting representative features using expert-designed descriptors or end-to-end deep neural networks, this paper is dedicated to discovering distinctive information from the shape formation process. Concretely, a spherical point cloud served as the template is progressively deformed to fit the target shape in a coarse-to-fine manner. During the shape formation process, several checkpoints are inserted to facilitate recording and investigating the intermediate stages. For each stage, the offset field is evaluated as a stage-aware description. The summation of the offsets throughout the shape formation process can completely define the target shape in terms of geometry. In this perspective, one can derive the point-wise shape correspondence from the template inexpensively, which benefits various graphic applications. In this paper, the Progressive Deformation-based Auto-Encoder (PDAE) is proposed to learn the stage-aware description through a coarse-to-fine shape fitting task. Experimental results show that the proposed PDAE has the ability to reconstruct 3D shapes with high fidelity, and consistent topology is preserved in the multi-stage deformation process. Additional applications based on the stage-aware description are performed, demonstrating its universality.

\end{abstract}

%
%
%
%
%
%

\section{Introduction}
Our key motivation comes from the sculpture creation process.
Imagine the carving process, in which the artist progressively removes materials from a stone to create 3D visual arts.
In different stages of carving, more and more details are embossed on top of low-fidelity surfaces. Therefore, one can intuitively assume that the distinctive information of the shape is progressively added throughout the sculpting. 
Fig. \ref{teaser} illustrates the shape formation process, in which a spherical point cloud is progressively deformed to fit the target shape.


The coarse-to-fine manner has been widely used and proven to be effective in various 3D computer graphics tasks, such as 3D shape recognition, segmentation \cite{qi2017pointnetpp, wang2019voxsegnet}, reconstruction \cite{wang2018pixel2mesh}, generation \cite{li_sig17_GRASS}, \etc.

For the recognition and segmentation problems, multi-scale features are extracted and integrated in a coarse-to-fine manner to enrich the descriptive ability for the encoding.
For the reconstruction and generation problems, through a coarse-to-fine manner, shape details can be added progressively in the generation process. The original problem can be decomposed into multiple sub-problems, which are simpler to solve. Recent works in both problems usually combine features from different scales to form a relatively comprehensive latent code for specific tasks \cite{li_sig17_GRASS, qi2017pointnetpp, wang2018pixel2mesh, wang2019voxsegnet}. In other words, they are task-driven methods.

Task-driven learning usually requires labour-intensive annotation such as object class label, point-wise semantic label, dense correspondence, \etc. 
The speed of knowledge enchanting is much slower than that of data increasing.
In addition to having high demand on data annotation, a task-driven method usually learns features that are highly relevant to the specific task, which decreases the universality of features.
Thus, in a big data surge background, semi-supervised, self-supervised, or unsupervised general purpose representation learning is required. The learned representation can further be used for various applications directly or with slight fine-tuning.

In this paper, a novel 3D shape representation is introduced, which is named as \textit{Cloud Sphere} according to the spherical point cloud template. We propose the Progressively Deformation-based Auto-Encoder (PDAE) which models the shape formation process.
A spherical point cloud served as the template is progressively deformed to fit the target shape in a coarse-to-fine manner. Several checkpoints are inserted to facilitate recording and investigating the intermediate stages. For each stage, the offset field is evaluated as a stage-aware description. The set of per-stage offsets throughout the shape formation process can completely define the target shape in terms of geometry. 
The Cloud Sphere representation explicitly encodes the deformation residuals during shape formation process. Using this representation, any shape can be modelled by a series of deformation fields from the spherical point cloud template. The same template acts as a bridge connecting different target shapes, intuitively provides their dense correspondences.


There are several differences between the proposed the Cloud Sphere representation and the task-driven ones. 
First, this representation is learnt through an unsupervised self-reconstruction framework. Therefore, additional annotations are not required.
Second, while learning the representation, multi-scale features act as conditions respectively in different reconstruction stages, rather than integrated as one latent code for a specific high-level 3D understanding task. This enables the learnt features to encode knowledge from corresponding reconstruction level. For interpretability, this method paves the way for studying the features' properties at different scales.
Third, the template deformation framework naturally provides point-wise correspondence between the template and the reconstructed point clouds. Note that point-wise correspondence is the low-level geometric information, which is of benefit to various high-level graphic applications. This is considered as the foundation of the universality of Cloud Sphere.

Our main contributions can be summarized as follows:

1) A novel 3D shape representation named Cloud Sphere which encodes knowledge from different stages of shape formation process.

2) A novel Progressively Deformation-based Auto-Encoder for point cloud feature encoding. This method enables the feature learning from each shape formation stage.


3) Experimental results demonstrate that the PDAE possesses better self-reconstruction ability compared to the
state-of-the-arts, and the correspondence obtained from the Cloud Sphere representation preserves better local and global topology.


4) The point-wise correspondence between different shapes can be obtained from the Cloud Sphere representation. Based on our method, various computer graphics applications are performed.

\section{Related Work}
In the area of 3D shape analysis, the geometric properties of a shape have long been studied.
The expert-designed descriptors \cite{BenChen2008, HWAG2009, Katz2003, Shapira2010, Zhang2012} focus on specific geometric properties. However, a single descriptor is often insufficient to cover complex high-level parsing scenarios. More detail discussion of these descriptors is given in \cite{Kalogerakis2010}.
Deep learning methods \cite{ Kalogerakis2017-ShapePFCN, qi2017pointnet, qi2017pointnetpp, yi2016syncspeccnn, Zhirong15CVPR} usually map the shape into a vector in high-dimensional space. While the descriptive capacity of features increases, the interpretability decreases. The interested readers are referred to an overview of deep geometry learning \cite{xiao2020survey}.


In this work, we use a progressive deformation framework for 3D shape self-reconstruction to learn the Cloud Sphere representation. Therefore, we mainly focus on the most related works which could be clustered into two classes: 1) multi-scale 3D shape representation and 2) deformation-based 3D shape modeling.

\subsection{Multi-Scale 3D Shape Representation}
Multi-scale features are widely used to enhance the features' descriptive ability and robustness.
Qin \etal proposed PointDAN \cite{qin2019pointdan}, a 3D Domain Adaptation Network for point cloud data. This method jointly aligns the global and local features at multiple levels.
Liu \etal proposed Point2Sequence \cite{liu2019point2sequence} to learn 3D shape features from point clouds. This method employs an attention-based sequence learning model to utilize a sequence of multi-scale areas of each local region. In which the correlation of different areas are captured to enhance the discriminability of learned features.
PointNet \cite{qi2017pointnet} operates on unordered point set directly. The independently processed points are aggregated into global feature by max-pooling.
In the follow-up work PointNet++ \cite{qi2017pointnetpp}, the authors improved PointNet by incorporating local dependencies and hierarchical feature learning in the network. This method is the first to extract features for multi-scale local regions individually and aggregates these features by concatenation, where the two steps are repeated to complete the hierarchical feature extraction.

The hierarchical multi-scale local region abstraction framework by PointNet++ is then followed by many works for point cloud understanding \cite{gadelha2018multiresolution,guo2020using,yin2019logan}.
For example, Guo \etal \cite{guo2020using} proposed a multi-scale deep network for semantic classification of terrestrial laser scanner (TLS) point clouds. This method subsamples the input point cloud at multiple scales, and then inputs each scale into the 3D CNN. The outputs are combined to form multi-scale and hierarchical deep features.
Gadelha \etal presented MRTNet \cite{gadelha2018multiresolution} (multi-resolution tree-structured networks) to process point clouds for 3D shape understanding and generation tasks. This method represents a 3D shape as a set of 1D ordered list of points.
Yin \etal introduced LOGAN \cite{yin2019logan}, which uses multi-scale features extracted from PointNet++ for unpaired 3D shape domain transfer. This method first embeds the shapes in a common latent space through a self-reconstruction task, and then trains an additional network for shape domain transfer. In the shape representation step, multi-scale features are concatenated as a comprehensive latent code.

\subsection{Deformation-Based 3D Shape modeling}
Deformation-based 3D modeling has been studied for a long time. Here we focus more on recent deep learning based methods. The deformation-based framework can be applied to various 3D data formats including mesh, voxel and point cloud. In the following, we give a fast review of these methods according to the data formats.

\textbf{Mesh data}. Mesh deformation has been used for 3D surface generation \cite{sinha2017surfnet}, single-view 3D mesh reconstruction \cite{wang2018pixel2mesh}, target-driven deformation \cite{wang20193dn}, \etc.
Specifically, Sinha \etal proposed SurfNet \cite{sinha2017surfnet} to use a parametrization of one or a few base shapes to represent training data as parametrized meshes. The vertex coordinates are mapped to the resulting UV space and fed into 2D neural networks for surface generation.
Wang \etal proposed Pixel2Mesh \cite{wang2018pixel2mesh}, a graph-based convolutional neural network which directly produces a 3D shape in triangular mesh from a single color image. An ellipsoid is progressively deformed and remeshed to fit the target 3D shape described in the 2D image.
Wang \etal introduced the 3D Deformation Network (3DN) \cite{wang20193dn}, an end-to-end network deforming a source 3D mesh to resemble a target shape in various kinds of modalities. They estimated the per-vertex offset displacements from the source mesh to the target shape. These methods usually require dense correspondence between the shapes \cite{sinha2017surfnet}, and generate results constrained by the input mesh topology.

\textbf{Voxel and point cloud data}. Voxel data can be seen as the regularized down-sampling of dense point clouds. Therefore we review these two data formats together.
Yumer \etal \cite{yumer2016learning} introduced the first 3D volumetric generative network that learns to predict per-voxel dense 3D deformation flows using explicit high level deformation intentions. This method uses Free Form Deformation (FFD) to apply the deformation flow to the original 3D shape.
Voxel-based methods usually have limited spatial resolution due to the high computational and memory cost.
Yang \etal proposed FoldingNet \cite{yang2018foldingnet}, which uses a folding-based decoding operation for point cloud reconstruction.
The folding operation that reconstructs a surface from a 2D grid essentially establishes a mapping from a 2D regular domain to a 3D point cloud.
In the same year, Groueix \etal introduced AtlasNet \cite{groueix2018papier}, a 3D shape surface generation framework. They represented 3D shape as a collection of parametric surface elements. The primitive surface elements are projected into 3D space to fit the target 3D shape.
Based on the idea of deforming 2D planes to 2-manifolds in 3D space used by FoldingNet and AtlasNet, Mehr \etal introduced DiscoNet \cite{mehr2019disconet} to learn a more complicated 3D shape embedding. Taking account of the topological variability in sets of 3D shapes, multiple branches of auto-encoders are trained to learn feature embeddings, which are not necessary to lie on a connected manifold. The decoding phase is performed by deforming a pre-learned 3D template representing the mean shape to the target instance.
Yin \etal introduced P2P-NET \cite{yin2018p2p}, a bi-directional point displacement network which learns geometric transformations between point clouds from two domains.
In summary, these methods perform point/voxel -wise deformation to generate new shapes.

In recent years, there emerge methods modeling the shape deformation process as a continuous differential problem \cite{jiang2020shapeflow, niemeyer2019occupancy, yang2019pointflow}. In these works, the continuous normalizing flow (CNF) \cite{chen2018neural, grathwohl2018ffjord} is used to establish a continuous invertible transformation with the help of a continuous-time dynamic. The CNF's outputs and the input gradients can be solved by ordinary differential equation (ODE) solvers. 
In this paper, we use a discrete deformation framework for the convenience to utilize multi-scale supervision.

\section{Definition of Cloud Sphere}

We propose Cloud Sphere, a novel 3D representation method for point clouds which explicitly encodes the deformation process during shape formation. This representation has some great properties for 3D shape analysis.
First, all shapes are modelled by a deformation from the spherical template, thus point-wise correspondences can be acquired effortlessly.
Second, point clouds are highly flexible. Without topological restrictions, the template is capable of fitting to arbitrary shapes with complex topological structure.
Third, through this representation, it is easy to model the hierarchical, progressive shape generation process, which is of benefit to high-level applications such as shape attribute localization and editing.

In the following, the Cloud Sphere representation is defined mathematically.
In the Cloud Sphere representation, a 3D shape $S:=\{s_{i} \in R^3\}_{i=1}^{n}$ is modelled by a spherical point cloud template  $T:=\{q_{i} \in R^3\}_{i=1}^{n}$ and the point-wise displacements $D^k:=\{d_{i}^{k} \in R^3\}_{i=1}^{n}$ from the sphere to the 3D shape surface. Note that $S$ and $T$ are point sets with the same cardinality, and $D$ is the point-wise offsets corresponding to the points in $T$. The stage of shape formation process is represented by $k$. The larger $k$ is, the higher abstract level of the shape is. The reconstructed shape of higher abstract level contains less geometric details.
The target shape $S$, spherical template $T$ and the point-wise displacements $D^k$ subject to the equation as:

\begin{equation}
   \centering
   S = T + \sum_{k=0}^{K}D^{k}.
   \label{cloud_sphere_def}
\end{equation}

Define $D:=\{D^k|k\in \{0,...,K\}\}$ as the set of the point-wise offsets in different shape formation stages. The deformation from the spherical point cloud $T$ to the target point cloud $S$ is decomposed into different deformation stages. The shape formation is the coarsest (close to $T$) while $k=K$, and is the finest (close to $S$) while $k=0$ . 
In consequence, the shape $S$ can be defined by $\{T, D\}$, which is named the Cloud Sphere representation.


This is a coarse-to-fine manner to add up a series of deformations $D^k$ together, which allows the shape to be gradually refined in detail. In addition, the multi-stage deformation process directly encodes distinctive information, which can help shape analysis and high-level applications.

\begin{figure*}[ht]
      \centering
      \includegraphics[width=1.0\linewidth]{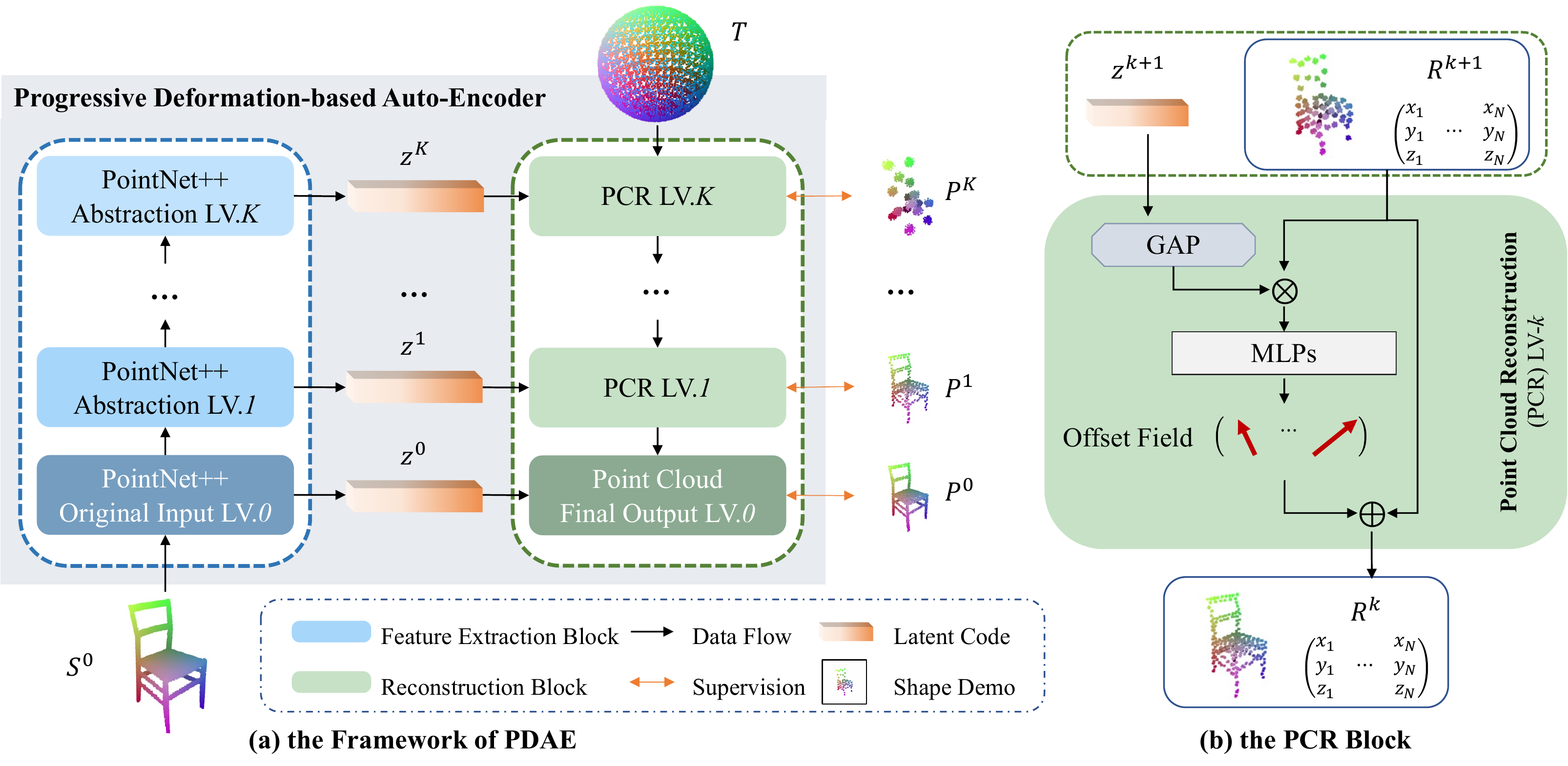}
      \caption{(a) The coarse-to-fine progressive deformation based auto-encoder framework is illustrated in the left block. Given shape $S^0$ as input, multi-level features are extracted by PointNet++. Then, the spherical template $T$ is progressively deformed to fit the multi-resolution target shapes $\{P^k\}_{k=0}^K$ from coarse to fine (from $K$ to $0$), through cascaded PCR blocks. (b) The detailed PCR block is shown in the right block. The PCR block takes the latent code and coarser level reconstruction as inputs, and predicts the point-wise offsets for finer-level reconstruction. We denote $\otimes$ as concatenation and $\oplus$ as point-wise summation, and global average pooling as GAP.}
      \label{framework}
\end{figure*}

\section{Method}

In this section, we describe the Progressive Deformation-based Auto-Encoder (PDAE) for the Cloud Sphere representation learning.
First, for computational feasibility, we model the ideal equation Eq. \ref{cloud_sphere_def} as a practical optimization problem.
Second, based on the practical problem, we introduce the network design for the PDAE.
Third, we give a detailed description about the main constraints to train the network.

\subsection{Problem modeling}
As defined in Eq. \ref{cloud_sphere_def}, a shape $S$ can be represented by the summation of a spherical template $T$ and the corresponding displacements $D$. In practice, it is not common that different shapes have the same point order in point cloud format; instead, the point set is naturally independent of the permutation order. Therefore, the equation $S = T + \sum_{k=0}^{K}D^k$ cannot hold in common. To make this problem computationally feasible, it is modelled as an optimization objective function:

\begin{equation}
\centering
\begin{split}
min_{D}\  dist(S, T+\sum_{k=0}^{K}(D^k)).
\end{split}
\label{cloud_sphere_objective}
\end{equation}

In this objective function, the distance between two point clouds are minimized. We choose the Chamfer Distance (CD) between two point sets as the measurement for shape similarity. Detailed description can be found in section \ref{constraints}.

\subsection{Network Design}

In order to solve the optimization objective defined in Eq. \ref{cloud_sphere_objective}, we design the Progressive Deformation-based Auto-Encoder, modeling the coarse-to-fine progressive shape formation process.

As shown in Fig. \ref{framework}, a 3D shape $S^0$ in point cloud format is fed into the encoding phase, which consists of the multi-scale feature extraction module. In this module, features from different levels of abstraction are extracted using the set abstraction modules from PointNet++ \cite{qi2017pointnetpp}. The abstracted feature from the top abstraction level $z^K$ is then fed into the corresponding coarsest (the $K_{th}$ level, in short, $LV.K$) Point Cloud Reconstruction (PCR) module, in which the spherical template is deformed to fit the coarsest level of reference point cloud $P^K$. Multiple PCRs in parallel with respect to the set abstraction modules are stacked to form the decoding phase. Through which the input spherical template is progressively deformed to reconstruct the input 3D shape.

The key variables in the PDAE can be formulated as follows:

\begin{equation}
\centering
\begin{split}
z^{k}, S^k &= f(z^{k-1}, S^{k-1}),\\
D^{k} &= g(z^{k}, R^{k+1}),\\
R^{k} &= R^{k+1} + D^{k},\\
\end{split}
\label{network_equations}
\end{equation}
where the superscript $k$ represents the level of abstraction. The larger $k$ is, the coarser shape is represented, and the less geometric details are depicted. Let $z^{k}$ denote the feature extracted from the point cloud in the abstraction level $k$. Let $S^{k}$ denote the point cloud in abstraction level $k$, which is sub-sampled $k$ times from the input. Let $D^{k}$ denote the predicted displacement field for shape reconstruction in level $k$, and $R^{k}$ denote the output point cloud of the reconstruction in level $k$. Particularly, $S^{0}$ is the original input point cloud, $T$ is the input spherical template. The spherical template is assigned to the initial reconstruction, i.e. $R^{K+1}=T$. The output $R^{0}$ is the finest level of reconstruction.
The set abstraction module from PointNet++ is formulated as $f(\cdot, \cdot)$, which takes the lower (finer) level latent code and point cloud as inputs, and outputs the current level latent code as well as the abstracted point set.
The point cloud reconstruction module is formulated as $g(\cdot, \cdot)$, which takes the higher (coarser) level reconstruction $R^{k+1}$ and the current level latent code $z^k$ as inputs, and outputs the current level reconstruction $R^k$.

After the PDAE is trained, the coarse-to-fine progressive representation Cloud Sphere can be extracted as:

\begin{equation}
\centering
CS := \{T, D^{k}, [z^{k}] | \{k=0,...,K\}\},
\label{cloud_sphere_extracted}
\end{equation}
in which the latent code $z^k$ is optional to enrich the descriptive ability, but does not influence the definition completeness.
In other words, the PDAE can be explained as a residual learning process for point-wise displacements to progressively fit the multi-resolution point clouds.
The training constraints are introduced in the next subsection.

\subsection{Multi-resolution Residual Learning}
\label{constraints}
\begin{figure}[!ht]
    \centering
    \includegraphics[width=1.0\linewidth]{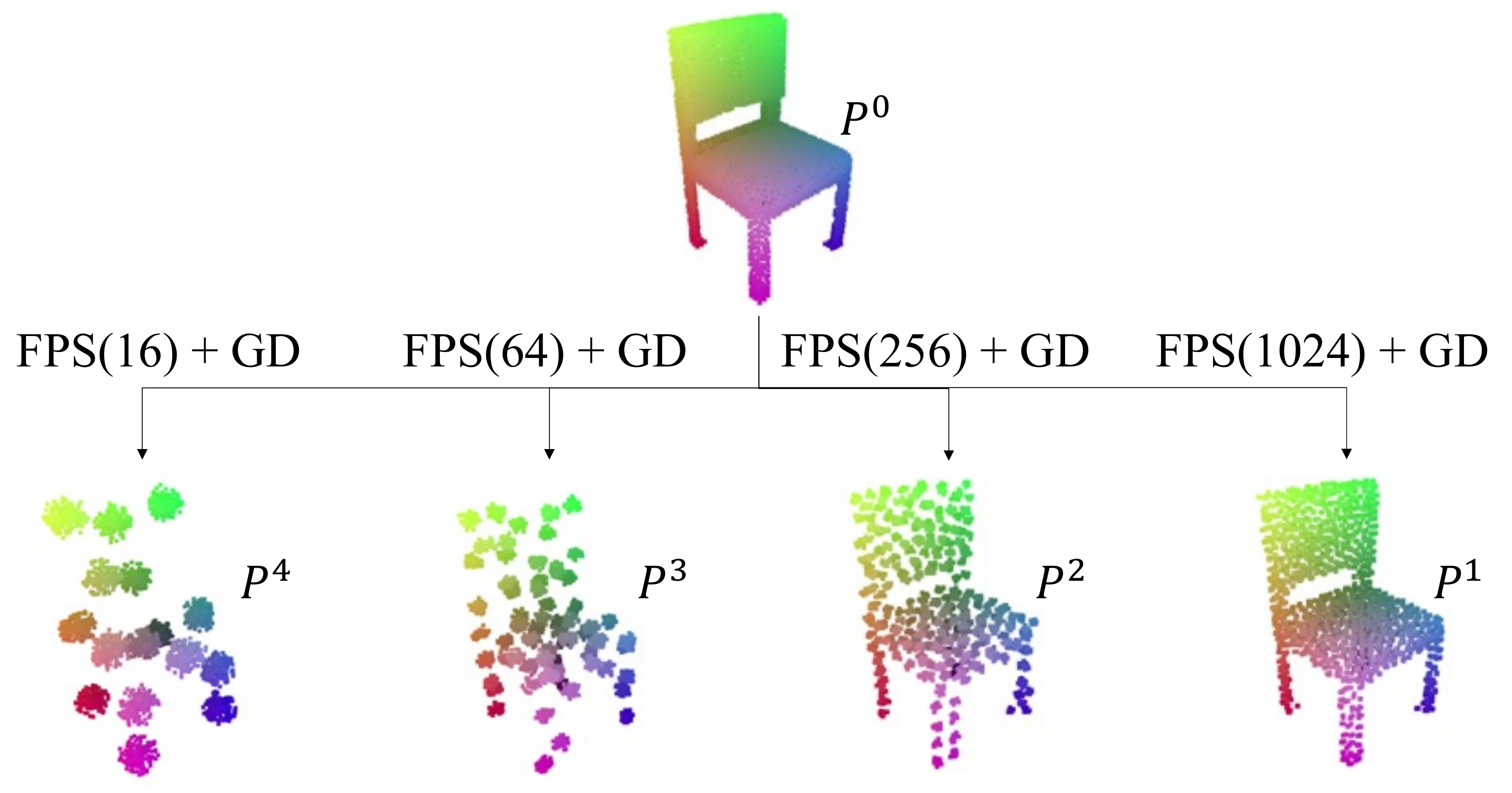}
    \caption{Supervision data preprocessing. The generated multi-resolution point clouds resemble the original point cloud from coarse to fine. FPS represents farthest points sampling. GD means Gaussian distribution.}
    \label{point_abs}
\end{figure}

\textbf{Multi-resolution data pre-processing.}
In order to guide the template-based progressive shape reconstruction task, the multi-resolution shapes from coarse to fine are required. This paper proposes a novel shape abstraction method to automatically extract multi-resolution point clouds from the original one. This would lower the demand for training data in this framework. The data preprocessing method is briefly introduced in below.

As shown in Fig. \ref{point_abs}, a 3D point cloud is resampled into different levels of abstraction. Specifically, Farthest Point Sampling (FPS) is applied first for downsampling to get the shape abstraction. Then, for each sampled point, a subset of points are randomly splattered around it with Gaussian distribution. In particular, compared with other downsampling methods, FPS provides the point cloud abstraction with better coverage of the original one. The splattered points according to the abstracted point cloud helps provide more precise supervision.

The multi-resolution point clouds have following properties:
\begin{equation}
\centering
\begin{split}
&|P^{k}|=|P^{k+1}|, \forall  0\leq k<K ,\\
&P^{k}=\bigcup_{i=1}^{N^k} \{X|gauss(x|\mu=p_i^{k}, \sigma^2)\},\\
\end{split}
\label{sample_and_splatter}
\end{equation}
where $P^k = \{p_i^{k} \in R^3\}_{i=1}^n$ represents the preprocessed result in abstraction level-$k$.

As revealed in the first equation in Eq. \ref{sample_and_splatter}, the generated point sets are with the same cardinality with the input, which helps the cascade of multi-stage deformation.
The second equation in Eq. \ref{sample_and_splatter} means that, the $k_{th}$ level resampled point cloud $P^k$ consists of several point subsets. Each of the subset is modeled by a set of random variable $\{X\}$ with $|\{X\}|=|P^k|/k$. The random variable $X$ obeys Gaussian distribution with the sampled point as mean, and $\sigma^2$ as the variation. The Gaussian distribution models the point existence probability. The closer to the sampled centroid, the more likely there exists points.

As a result, the generated multi-resolution point clouds resemble the original point cloud from coarse to fine. After point set downsampling and points splattering using Gaussian distribution, the high-level abstracted point clouds can cover low-level ones, but depict more about the shape structure. In the high-level abstraction stage with a relatively low sample-rate, compared to point clouds with merely downsampling, the augmented ones with splattered points benefit from better balance for the distance metrics.

\textbf{Shape similarity constraint.}
We use Chamfer Distance to measure the difference between two point sets:

\begin{equation}
\centering
\begin{split}
L_{CD}(k) = &\sum_{p\in P^{k}}\min_{q\in R^{k}}||p - q||_2^2 +
         \sum_{q\in R^{k}}\min_{p\in P^{k}}||q - p||_2^2 , \\
\end{split}
\label{recon_loss}
\end{equation}
where $k$ is the level of abstraction. This Chamfer Distance is invariant to the ordering of the output point cloud. The multi-stage shape similarity loss can be written as:

\begin{equation}
    \centering
    \begin{split}
        L_{CD} = \sum_{k=0}^K \alpha_k L_{CD}(k),
    \end{split}
    \label{recon_loss_total}
\end{equation}
where $\alpha_k$ represents the weight of the shape similarity on the abstraction level-$k$.

\textbf{Point set topology regularization.}
To regularize the point-wise deformation, we use the local topology preserving term to encourage the deformed template to preserve its own topology.

\begin{equation}
\centering
\begin{split}
\omega_{i,j}=exp(-||q_i - q_j||_2) , \forall q_{*} \in T, \\
L_{reg}(k) = \sum_{0<i\neq j\leq N}\omega_{i,j}||d^k(i) - d^k(j)||_2,
\end{split}
\label{topo_regularize}
\end{equation}
in which $\omega_{i,j}$ is the pairwise weight between two points $(q_i, q_j)$ from the spherical template $T$. The closer they are, the bigger $\omega_{i,j}$ is. Let $k$ denote the shape abstraction level, therefore $d^k(i)$ is the displacement of the point $q_i$ in $R^{k+1}$. Note that the displacement vector $d^k$ has the same order with the spherical template $T$. The multi-stage topology constraints can be written as:

\begin{equation}
    \centering
    \begin{split}
        L_{reg} = \sum_{k=0}^K \beta_k L_{reg}(k),
    \end{split}
    \label{topo_regularize_total}
\end{equation}
in which $\beta_k$ represents the weight of the topology regularization on the abstraction level-$k$.

The total loss $L_{total}$ is:
\begin{equation}
\centering
\begin{split}
L_{total} = \sum_{k=K}^0 (\alpha_{k}L_{CD}(k) + \beta_{k}L_{reg}(k)).
\end{split}
\label{loss_total}
\end{equation}
We set the weights as $\alpha|_k=[0.5,0.2,0.2,0.2,0.2]$ and $\beta|_k=[0,0,0,1.0,10.0]$ with $k=[0,1,2,3,4]$.

\section{Experimental Results}
\label{sec:experimental_results}
In this section, we first present the data we used to train and test the proposed PDAE framework.
Two experiments are conducted to evaluate the performance of the PDAE. First, self-reconstruction results are qualitatively and quantitatively compared with the state-of-the-art methods. Second, the correspondence results achieved by deformation-based reconstruction are evaluated. Next, an ablation study is conducted to verify the effectiveness of the progressive deformation framework. Finally, several interesting applications based on the Cloud Sphere representation are demonstrated.

\subsection{Dataset}
Our data is rearranged from the standard ShapeNet Core dataset (v1) \cite{chang2015shapenet}. Specifically, we select 4 major semantic sets containing more than 1000 instances in total: airplane, car, chair and table. For each semantic set, we analyse the fine-grained classes. Finally, 4 different fine-grained classes from each semantic class are achieved. After the rearrangement of the ShapeNet Core (v1) dataset, we obtain a challenging dataset for fine-grained shape analysis, on which generalized correspondence among shapes within a coarse class can be defined. The point clouds are generated using the virtual scanner from O-CNN \cite{wang2017cnn}, and then for each shape, 4096 points are sampled using FPS.

\begin{figure*}[!ht]
    \centering
    \includegraphics[width=1.0\linewidth]{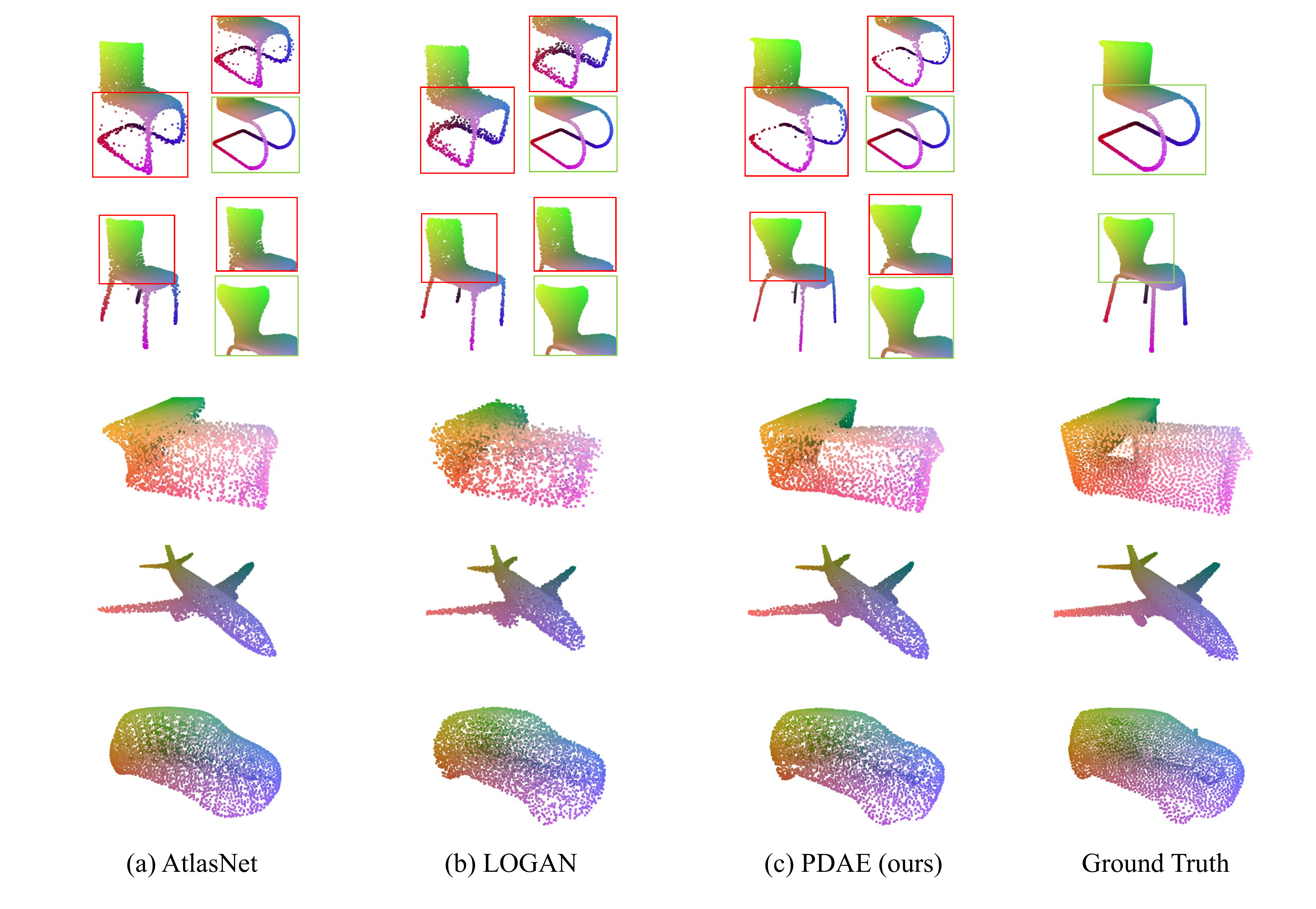}
    \caption{Reconstruction comparison with two methods. The proposed PDAE produces shapes with more geometric details and less noisy points. In the top two rows, the comparison areas are highlighted with boxes. The red boxes represent the output results, while the green boxes represent the ground-truth comparison.}
    \label{recon_comparison}
\end{figure*}

\subsection{Self-reconstruction Evaluation}
We compare the proposed PDAE with LOGAN \cite{yin2019logan} and AtlasNet \cite{groueix2018papier} on the shape self-reconstruction task. In AtlasNet, the same spherical point cloud template as ours is used as input. Table \ref{table_self_reconst} shows the quantitative results. Our method produces point clouds reconstruction with the lowest Chamfer distance and earth mover distance, and highest intersection over union on solid voxelized grids.

Fig. \ref{recon_comparison} illustrates the qualitative results. The PDAE produces results with more accurate geometry and topology. We can see in the top two rows, the chair leg and back from the PDAE have more geometric details and less noisy points. Similar effects can be observed in the other examples including the table (the third row), the plane (the forth row) and the car (the bottom row).

\begin{table}[ht]  
   \small
   \renewcommand{\arraystretch}{1.2}
   \caption{Self-reconstruction results on test sets. Chamfer Distance (CD $\times 1000$), Earth Mover Distance (EMD $\times 100$), and Intersection over Union (IoU) on solid voxelized grids are estimated. The $\downarrow$ means the lower the better. The $\uparrow$ means the higher the better.
   }
   \label{table_self_reconst}
   
   \centering
   \resizebox{0.5\textwidth}{!}{
       \begin{tabular}{L{0mm} l C{3mm}C{3mm}C{3mm} C{3mm}C{3mm}C{3mm} C{3mm}C{3mm}C{3mm}}
           \toprule
           &\ &\multicolumn{3}{c}{LOGAN \cite{yin2019logan}} &\multicolumn{3}{c}{AtlasNet \cite{groueix2018papier}} &\multicolumn{3}{c}{PDAE (ours)} \\
           \cmidrule(lr){3-5} \cmidrule(lr){6-8} \cmidrule(lr){9-11}
           &Metrics  & ~CD $\downarrow$  & EMD $\downarrow$ & ~IoU $\uparrow$  & ~CD $\downarrow$ & EMD $\downarrow$ &~IOU $\uparrow$ & ~CD $\downarrow$ & EMD $\downarrow$ & ~IoU $\uparrow$\\
           \midrule
           &Plane A 
           &0.23	&1.07	&0.77	
           &0.17	&0.94	&0.81	
           &0.15	&0.90	&0.81
           \\
           
           &Plane B 
           &0.82	&1.93	&0.61	
           &0.64	&1.67	&0.62	
           &0.40	&1.43	&0.69
           \\
           
           &Plane C  
           &3.20	&3.73	   &0.40	
           &1.31	&2.39	   &0.55	
           &0.68	&1.98    &0.61
           \\
           
           &Plane D
           &1.60	&3.18	&0.48	
           &0.92	&2.29	&0.51	
           &0.78   &2.30	&0.53
           \\
           
           &Mean  
           &1.46	&2.48	&0.56	
           &0.76	&1.82	&0.62	
           &\textbf{0.50}	&\textbf{1.65}	&\textbf{0.66}
           \\
           
           \midrule
           &Car A 
           &0.62	&1.83   &0.56	
           &0.70	&1.57	&0.66	
           &0.40	&1.53	&0.67
           \\
           
           &Car B 
           &0.37	&1.50	&0.64	
           &0.85   &1.82	&0.61	
           &0.33	&1.42	&0.69
           \\
           
           &Car C  
           &0.99	&2.33	&0.45	
           &7.46	&3.52   &0.51	
           &0.62	&1.82	&0.59
           \\
           
           &Car D
           &0.76	&2.04	&0.54	
           &0.76	&1.91	&0.55	
           &0.47	&1.69	&0.63
           \\
           
           &Mean  
           &0.69	&1.92	&0.55	
           &2.44	&2.21	&0.58	
           &\textbf{0.45}	&\textbf{1.62}	&\textbf{0.65}
           \\
           
           \midrule
           &Chair A 
           &1.59	&3.16	&0.46	
           &0.92	&2.43	&0.54	
           &0.51	&1.68	&0.71
           \\
           
           &Chair B 
           &5.04	&5.15	&0.32	
           &2.00	&3.43	&0.43	
           &1.41	&2.71	&0.56
           \\
           
           &Chair C  
           &4.61	&5.00	&0.29	
           &2.09	&3.35	&0.43	
           &1.31	&2.69	&0.59
           \\
           
           &Chair D
           &4.50	&5.18	&0.32	
           &1.86	&2.97	&0.49	
           &1.66	&2.93	&0.52
           \\
           
           &Mean  
           &3.94	&4.62	&0.35	
           &1.72	&3.04	&0.47	
           &\textbf{1.22}	&\textbf{2.50}	&\textbf{0.59}
           \\
           
           \midrule
           &Table A 
           &7.41	&6.34	&0.25	
           &2.11	&3.38	&0.41	
           &1.64	&2.90	&0.50
           \\
           
           &Table B 
           &5.38	&5.64	&0.25	
           &2.22	&3.54	&0.36	
           &1.05	&2.41	&0.55
           \\
           
           &Table C  
           &1.57	&2.92	&0.41	
           &2.21	&2.88	&0.43	
           &0.53	&1.84	&0.67
           \\
           
           &Table D
           &4.44	&4.96	&0.32	
           &2.56	&3.56	&0.40	
           &0.89	&2.17	&0.62
           \\
           
           &Mean  
           &4.70	&4.96	&0.31	
           &2.28	&3.34	&0.40	
           &\textbf{1.03}	&\textbf{2.33}	&\textbf{0.59}
           \\
           
           \bottomrule
       \end{tabular}
   }
\end{table}

\subsection{Correspondence Evaluation}
\begin{figure}[!ht]
\centering
\includegraphics[width=0.95\linewidth]{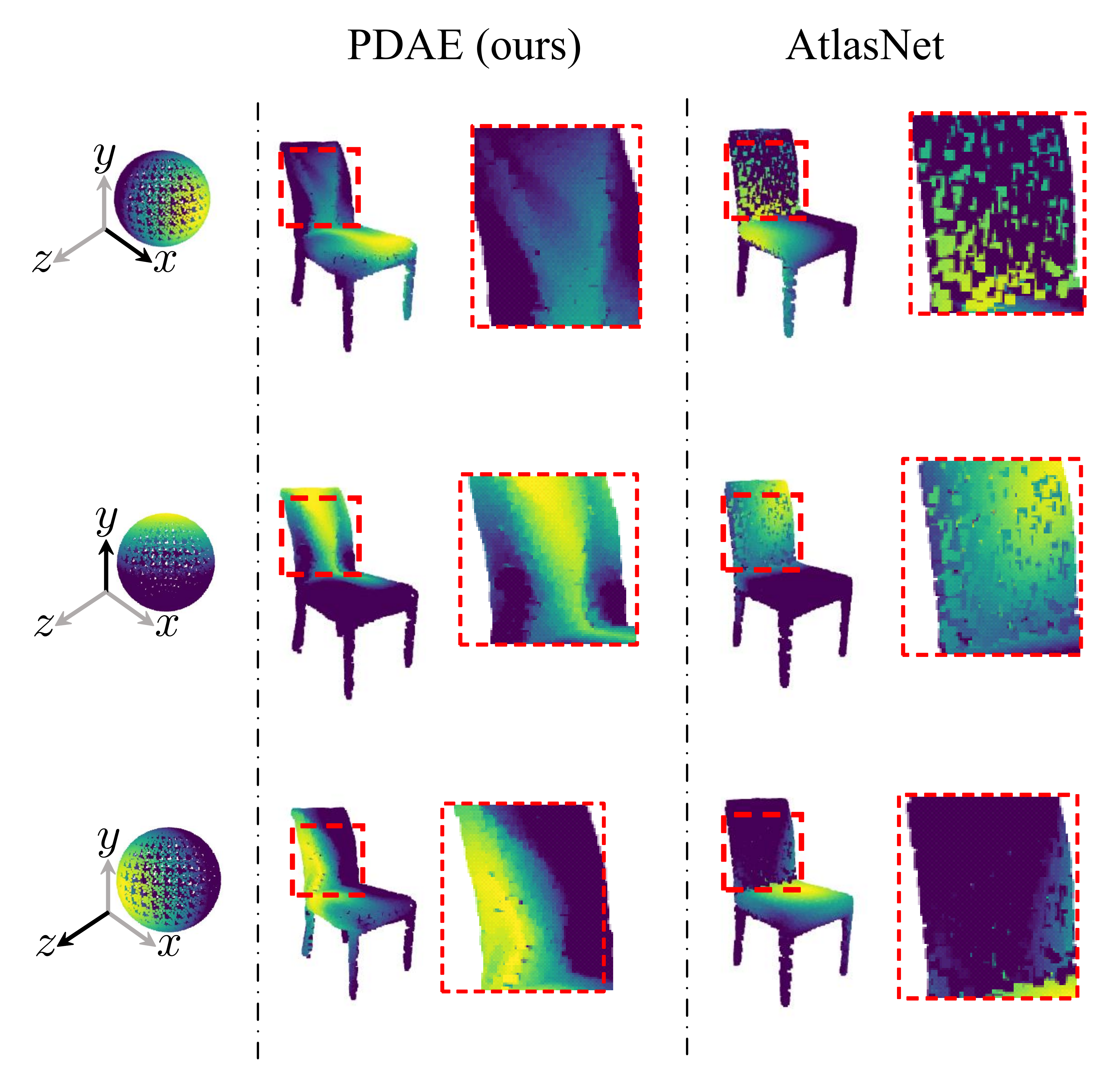}
\caption{Correspondence comparison between our method and AtlasNet. In AtlasNet, the same spherical point cloud template as ours is used as input. The colors in the spherical template are transferred into the reconstructed chairs. It can be observed that our correspondence derived from PDAE preserves better local and global topology from the template.}
\label{correspondence_comparison}
\end{figure}

In a deformation-based reconstruction framework, the template is deformed to fit the target shape. Therefore, there exists point-wise correspondence between the original template and the reconstructed shape. We obtain the bi-projection between the spherical template $T$ and the reconstructed point cloud $R^{0}$. The spherical template plays the role of a bridge through which various shapes are connected. In this experiment, we evaluate the template-to-shape correspondence.

In Fig. \ref{correspondence_comparison}, the template sphere is color-coded according to the coordinate values along three axes respectively. The color is then directly projected to the reconstructed shapes. For each row in Fig. \ref{correspondence_comparison}, the left column is the color coding direction, the middle is the correspondence result from our PDAE, and the right is that from AtlasNet. 
We observed that in the rightmost column (AtlasNet results), the color in the chair back is discontinued. Which implies that the near region on the chair back is constructed by faraway points from the template.
Moreover, from all the three rows, we can see the middle column results are with the same color direction ($x$,$y$,or $z$) as the template, while the right column results are with different color direction. This is because our method progressively defroms the template to fit the target, while AtlasNet directly projects the point's latent code into 3D space, without coarse-to-fine constraints.

We also provide the numerical results in Table \ref{table_correspondence}.
Considering the absence of ground-truth dense correspondence, we use $\overline{spread}$ to represent the extent of reconstruction distraction among the nearby points grouped by voxelized grids. Specifically, $spread$ is the variance of the reconstructed point locations indexed from the same voxelized grid in the template. $\overline{spread}$ is the average of $spread$ throughout the shape. The larger $\overline{spread}$ is, the messier template points are used to reconstruct the shape. 
We use $\overline{shift}$ to measure the average displacements from the template to the target shape. Our method achieves lower mean $spread$ and $shift$, which suggests that the correspondence obtained by PDAE preserves better local and global topology.

\begin{table}[ht]  
   \small
   \renewcommand{\arraystretch}{1.2}
   \caption{Correspondence evaluation. We denote $\overline{spread}$ as the extent of reconstruction distraction from the template point clouds. We use $\overline{shift}$ to measure the average displacements from the template to the target. For both metrics, the lower the better.}
   \label{table_correspondence}
   
   \centering
   \resizebox{0.40\textwidth}{!}{
       \begin{tabular}{L{0mm} l C{8mm}C{8mm} C{8mm}C{8mm}}
           \toprule
           &\ &\multicolumn{2}{c}{AtlasNet \cite{groueix2018papier}} &\multicolumn{2}{c}{PDAE (ours)} \\
           \cmidrule(lr){3-4} \cmidrule(lr){5-6}
           &Metrics  & $\overline{spread}\downarrow$  & $\overline{shift}\downarrow$  & $\overline{spread}\downarrow$  & $\overline{shift}\downarrow$\\
           \midrule
           &Plane A 
           &0.40	&1.21	&0.29	&0.79
           \\
           
           &Car A 
           &0.17	&1.13	&0.19	&0.61
           \\
           
           &Chair A  
           &0.48	&0.98	&0.23	&0.68
           \\
           
           &Table A
           &0.30	&1.07	&0.16	&0.60
           \\
           \midrule
           &Mean
           &0.34	&1.10	&\textbf{0.22} &\textbf{0.67}
           \\
           \bottomrule
       \end{tabular}
   }
\end{table}

\subsection{Ablation Study}
In order to verify the effects of the multi-stage supervision on feature learning, the ablation study on the number of deformation stages is conducted. The experiment is carried out in the Chair A dataset. Different number of supervision stages are used for training the network. Detailed settings are as follows:

1) Let $\{\}$ represent the training without multi-stage supervision. Only the features extracted from the original shape is used for deformation-based shape reconstruction.

2) Let $\{16\}$ represent the two-stage supervised training. In this setting, the coarsest shape $P^4$ is used as the intermediate supervision for the progressive deformation. The spherical template is deformed to fit the abstract coarse shape first, and then to fit the original shape.

3) Let $\{16, 256\}$ represent the three-stage supervised training. In this setting, the shapes $P^4$ and $P^2$ are used as the intermediate supervision for the progressive deformation.

4) Let $\{16, 64, 256, 1024\}$ represent the multi-stage supervision used in the proposed PDAE framework. In this setting, the shapes $P^4, P^3, P^2$ and $P^1$ are used as the intermediate supervision for the progressive deformation.

As is shown in Table \ref{table_stages}, the reconstruction task benefits from the proposed multi-stage supervision. It can be observed that, with the number of stages increases, there are better numerical evaluations of the reconstruction results.

\begin{table}[ht]  
    \small
    \renewcommand{\arraystretch}{1.2}
    \caption{Ablation study on the number of deformation stages. Chamfer Distance (CD $\times 1000$), Earth Mover Distance (EMD $\times 100$), and Intersection over Union (IoU) on solid voxelized grids are estimated.}
    \label{table_stages}
    
    \centering
    \resizebox{0.45\textwidth}{!}{
        \begin{tabular}{l c c c c}
            \toprule
            &Metrics  & CD$\downarrow$  & EMD$\downarrow$  & IoU$\uparrow$ \\
            \midrule
            &$\{\}$ &1.52	&2.73	&0.48
            \\
            
            &$\{16\}$ &0.96	&2.47	&0.57
            \\
            
            &$\{16,256\}$ &0.76	&2.06	&0.63
            \\
            
            &$\{16,64,256,1024\}$ &0.51	&1.68	&0.71
            \\
            \bottomrule
        \end{tabular}
    }
\end{table}

\subsection{Additional Applications}

\subsubsection{Class-defined Region Localization}

The Cloud Sphere representation can help localize the class-defined regions. Specifically, the Cloud Sphere of the 3D shape is fed into a PointNet \cite{qi2017pointnet} like encoder, through which multi-level features are extracted and concatenated into a comprehensive one. The combined feature is then passed through an squeeze and excitation module \cite{hu2018squeeze}. Finally a fully-connected layer is used to perform classification.

\begin{figure}[!t]
\centering
\includegraphics[width=1.0\linewidth]{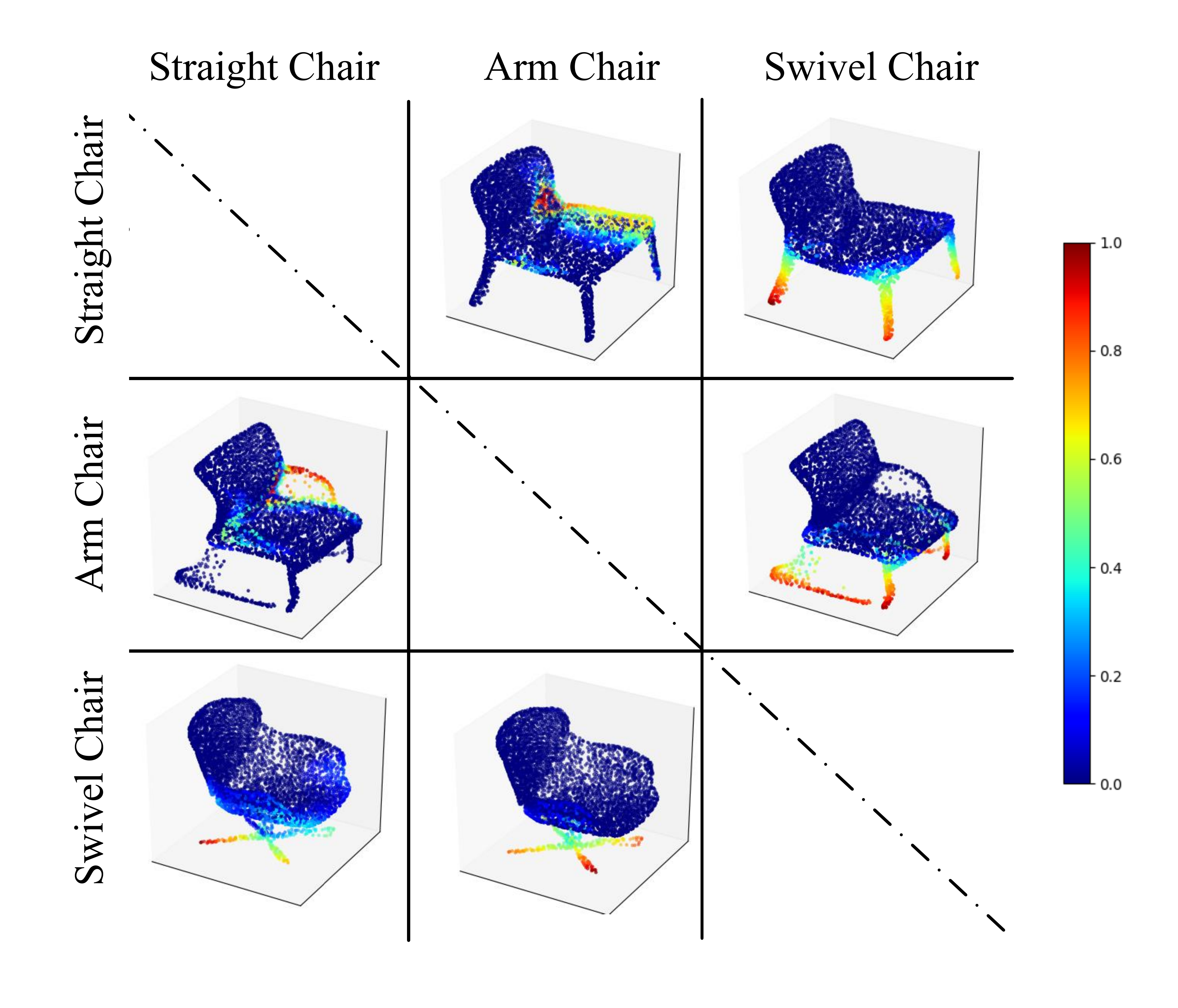}
\caption{Class-defined region localization. Three fine-grained classes of chairs are illustrated. The element $(i,j)$ represents the distinctive regions which are highlighted in shape $i$ compared to class $j$. This can be considered as the localization of the high-level semantic regions.}
\label{app_region_cls}
\end{figure}

Fig. \ref{app_region_cls} shows the class-defined region localization obtained by the classification task. The regions are detected by the point-wise activation in the high attention feature channel. Three fine-grained classes of chairs are illustrated. We highlight the most distinctive regions compared to each class.
This will be great convenient for multiple high-level graphic tasks such as style localization, shape transfer, \etc.

\subsubsection{Deformation-based Shape Editing}
\begin{figure}[!ht]
    \centering
    \includegraphics[width=1.0\linewidth]{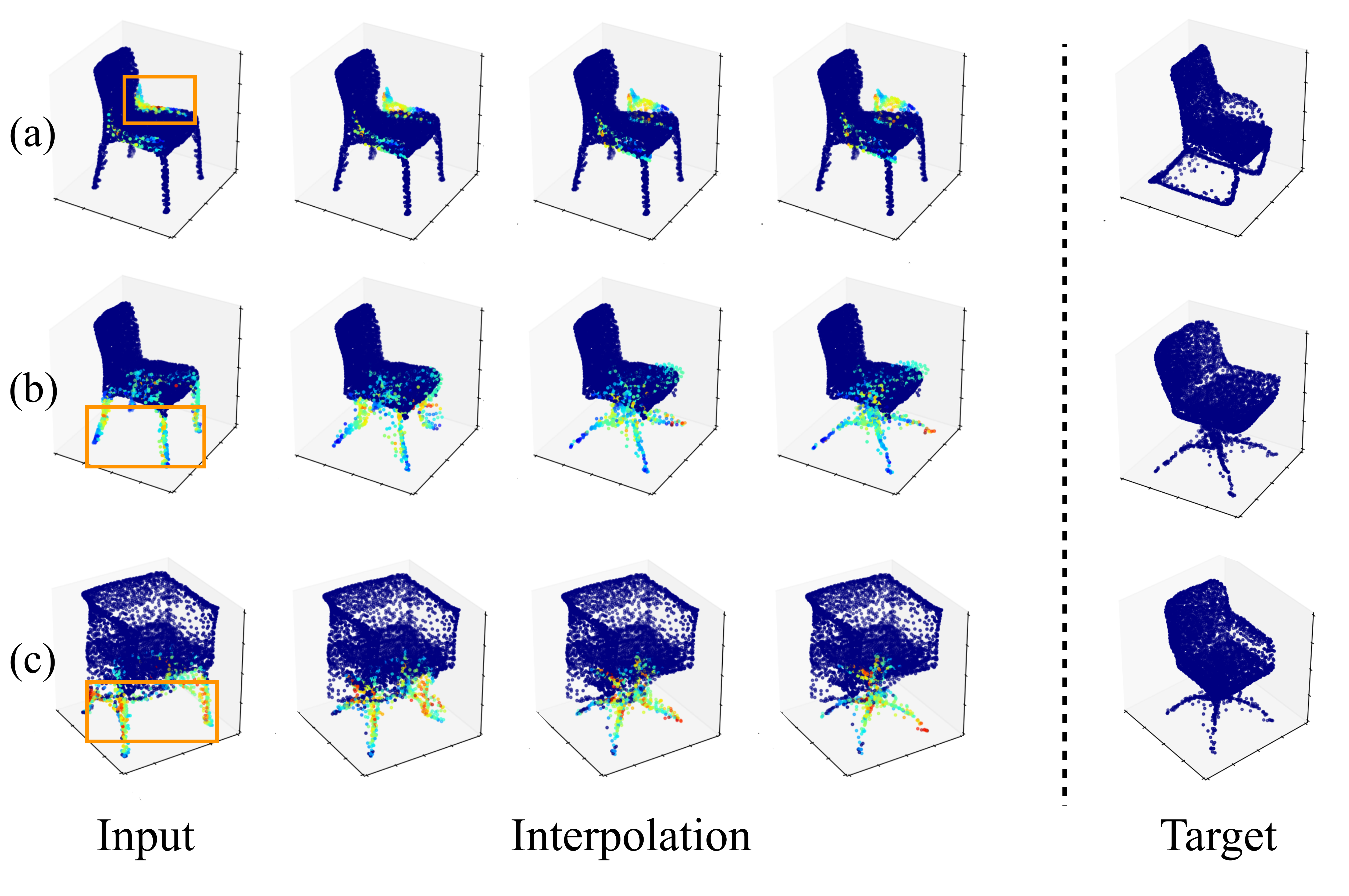}
    \caption{Shape geometry transfer editing results. The class-defined regions are highlighted in boxes and deformed to the target's counterpart. (a) The chair arms are lifted from the seat. (b) and (c) the chairs' straight legs are deformed to star-like ones.}
    \label{shape_transfer}
\end{figure}

\textbf{Shape Geometry Transfer.} This application is based on the inter-class correspondence. We perform shape geometry transfer between different fine-grained classes. Once the distinctive regions are detected, shape interpolation is then conducted based on the detection results. Specifically, only the class-distinctive regions are deformed towards the target. The remaining parts are maintained. Through this manner, fine-grained class or style transfer can be achieved. 

The geometry transfer results are illustrated in Fig. \ref{shape_transfer}. We can see arms are added to the proper position of the straight chairs. Note that the straight legs can be deformed to star-like ones.

\begin{figure}[!ht]
\centering
\includegraphics[width=0.95\linewidth]{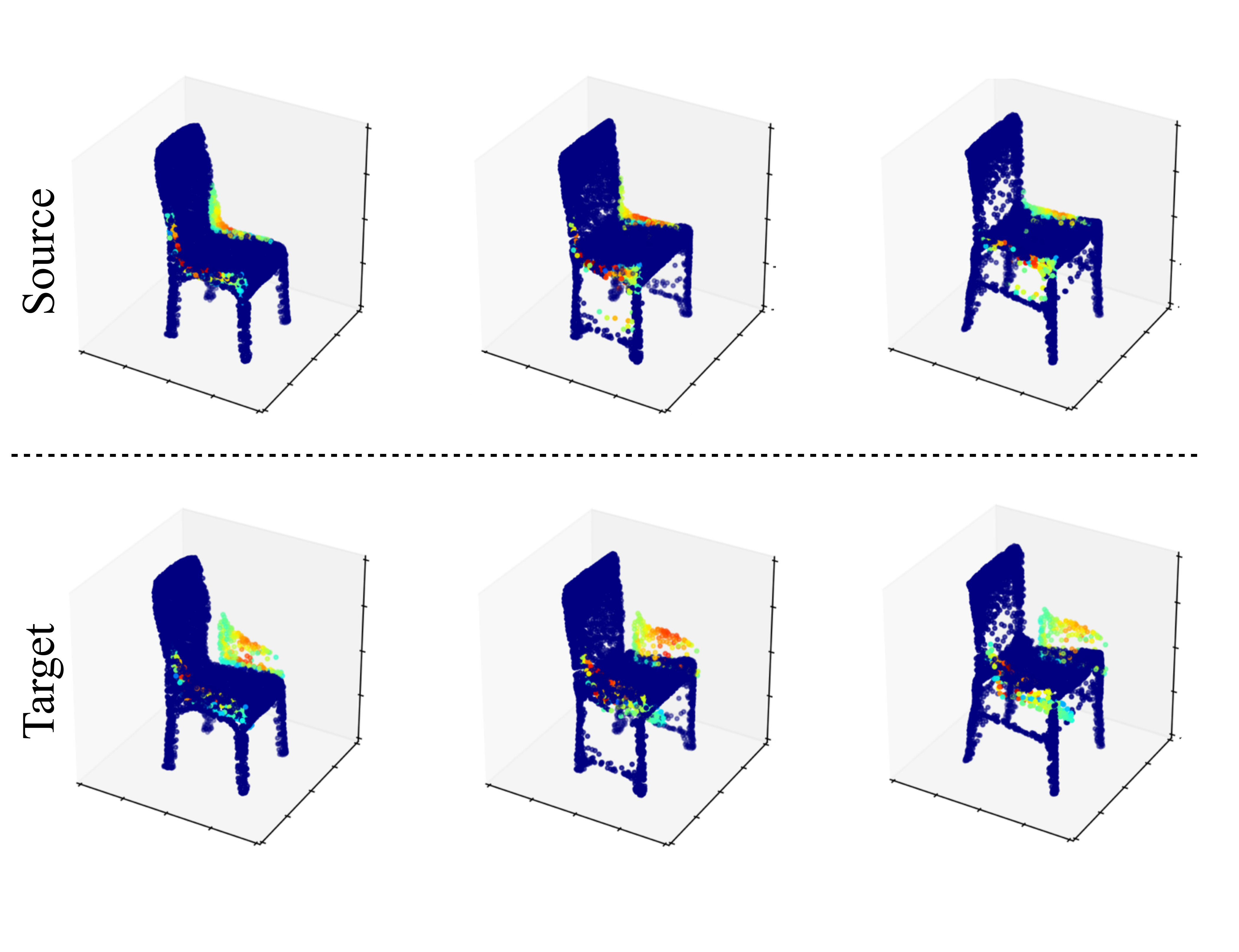}
\caption{Shape co-editing results. The arms are simultaneously added to the straight chair in the top row by deforming the highlighted class-defined region.}
\label{shape_co_editing}
\end{figure}

\noindent\textbf{Co-editing.} Using the inner-class correspondence, we can modify the shapes in the same class simultaneously. Fig. \ref{shape_co_editing} illustrates the co-editing results. The arms are added to a set of straight chairs simultaneously.



\section{Conclusion}

This paper has introduced a novel 3D shape representation method which explicitly encodes the deformation residuals from a spherical surface. The Cloud Sphere representation constructs point-wise correspondence between the shape and the template. Therefore it can serve as a general-purpose representation for various graphic applications. In order to learn this representation, The progressive deformation-based auto-encoder has been proposed. Specifically, a spherical point cloud is deformed progressively from coarse to fine to fit the multi-scale counterparts of the target shape, with the corresponding multi-scale features as guidance. Experimental results show that the proposed PDAE has the ability to reconstruct 3D shapes with high fidelity, and consistent topology is preserved while deformation. Additional applications based on the Cloud Sphere representation have been performed, demonstrating its universality.
In future, we plan to extend the multi-stage supervision to the continuous deformation framework.









{\small
\bibliographystyle{cvm}
\bibliography{egbib}

\begin{thebibliography}{10}\itemsep=-1pt

\bibitem{BenChen2008}
M.~Ben-Chen and C.~Gotsman.
\newblock Characterizing shape using conformal factors.
\newblock In {\em Proceedings of the 1st Eurographics Conference on 3D Object
  Retrieval}, 3DOR '08, pages 1--8, Aire-la-Ville, Switzerland, Switzerland,
  2008. Eurographics Association.

\bibitem{chang2015shapenet}
A.~X. Chang, T.~Funkhouser, L.~Guibas, P.~Hanrahan, Q.~Huang, Z.~Li,
  S.~Savarese, M.~Savva, S.~Song, H.~Su, et~al.
\newblock Shapenet: An information-rich 3d model repository.
\newblock {\em arXiv preprint arXiv:1512.03012}, 2015.

\bibitem{chen2018neural}
R.~T. Chen, Y.~Rubanova, J.~Bettencourt, and D.~Duvenaud.
\newblock Neural ordinary differential equations.
\newblock {\em arXiv preprint arXiv:1806.07366}, 2018.

\bibitem{gadelha2018multiresolution}
M.~Gadelha, R.~Wang, and S.~Maji.
\newblock Multiresolution tree networks for 3d point cloud processing.
\newblock In {\em Proceedings of the European Conference on Computer Vision
  (ECCV)}, pages 103--118, 2018.

\bibitem{grathwohl2018ffjord}
W.~Grathwohl, R.~T. Chen, J.~Bettencourt, I.~Sutskever, and D.~Duvenaud.
\newblock Ffjord: Free-form continuous dynamics for scalable reversible
  generative models.
\newblock {\em arXiv preprint arXiv:1810.01367}, 2018.

\bibitem{groueix2018papier}
T.~Groueix, M.~Fisher, V.~G. Kim, B.~C. Russell, and M.~Aubry.
\newblock A papier-m{\^a}ch{\'e} approach to learning 3d surface generation.
\newblock In {\em Proceedings of the IEEE conference on computer vision and
  pattern recognition}, pages 216--224, 2018.

\bibitem{guo2020using}
Z.~Guo and C.-C. Feng.
\newblock Using multi-scale and hierarchical deep convolutional features for 3d
  semantic classification of tls point clouds.
\newblock {\em International Journal of Geographical Information Science},
  34(4):661--680, 2020.

\bibitem{hu2018squeeze}
J.~Hu, L.~Shen, and G.~Sun.
\newblock Squeeze-and-excitation networks.
\newblock In {\em Proceedings of the IEEE conference on computer vision and
  pattern recognition}, pages 7132--7141, 2018.

\bibitem{HWAG2009}
Q.~Huang, M.~Wicke, B.~Adams, and L.~Guibas.
\newblock Shape decomposition using modal analysis.
\newblock In {\em Computer Graphics Forum (Proceedings of Eurographics 2009)},
  volume~28, pages 407--416, Munchen, Germany, April 2009.

\bibitem{jiang2020shapeflow}
C.~Jiang, J.~Huang, A.~Tagliasacchi, L.~Guibas, et~al.
\newblock Shapeflow: Learnable deformations among 3d shapes.
\newblock {\em arXiv preprint arXiv:2006.07982}, 2020.

\bibitem{Kalogerakis2017-ShapePFCN}
E.~Kalogerakis, M.~Averkiou, S.~Maji, and S.~Chaudhuri.
\newblock 3{D} shape segmentation with projective convolutional networks.
\newblock In {\em Proc. IEEE Computer Vision and Pattern Recognition (CVPR)},
  2017.

\bibitem{Kalogerakis2010}
E.~Kalogerakis, A.~Hertzmann, and K.~Singh.
\newblock Learning 3d mesh segmentation and labeling.
\newblock {\em ACM Trans. Graph.}, 29(4):102:1--102:12, July 2010.

\bibitem{Katz2003}
S.~Katz and A.~Tal.
\newblock Hierarchical mesh decomposition using fuzzy clustering and cuts.
\newblock {\em ACM Trans. Graph.}, 22(3):954--961, July 2003.

\bibitem{li_sig17_GRASS}
J.~Li, K.~Xu, S.~Chaudhuri, E.~Yumer, H.~Zhang, and L.~Guibas.
\newblock Grass: Generative recursive autoencoders for shape structures.
\newblock {\em ACM Transactions on Graphics (Proc. of SIGGRAPH 2017)}, 36(4):to
  appear, 2017.

\bibitem{liu2019point2sequence}
X.~Liu, Z.~Han, Y.-S. Liu, and M.~Zwicker.
\newblock Point2sequence: Learning the shape representation of 3d point clouds
  with an attention-based sequence to sequence network.
\newblock In {\em Proceedings of the AAAI Conference on Artificial
  Intelligence}, volume~33, pages 8778--8785, 2019.

\bibitem{mehr2019disconet}
E.~Mehr, A.~Jourdan, N.~Thome, M.~Cord, and V.~Guitteny.
\newblock Disconet: Shapes learning on disconnected manifolds for 3d editing.
\newblock In {\em Proceedings of the IEEE International Conference on Computer
  Vision}, pages 3474--3483, 2019.

\bibitem{niemeyer2019occupancy}
M.~Niemeyer, L.~Mescheder, M.~Oechsle, and A.~Geiger.
\newblock Occupancy flow: 4d reconstruction by learning particle dynamics.
\newblock In {\em Proceedings of the IEEE/CVF International Conference on
  Computer Vision}, pages 5379--5389, 2019.

\bibitem{qi2017pointnet}
C.~R. Qi, H.~Su, K.~Mo, and L.~J. Guibas.
\newblock Pointnet: Deep learning on point sets for 3d classification and
  segmentation.
\newblock In {\em Proceedings of the IEEE conference on computer vision and
  pattern recognition}, pages 652--660, 2017.

\bibitem{qi2017pointnetpp}
C.~R. Qi, L.~Yi, H.~Su, and L.~J. Guibas.
\newblock Pointnet++: Deep hierarchical feature learning on point sets in a
  metric space.
\newblock In {\em Advances in neural information processing systems}, pages
  5099--5108, 2017.

\bibitem{qin2019pointdan}
C.~Qin, H.~You, L.~Wang, C.-C.~J. Kuo, and Y.~Fu.
\newblock Pointdan: A multi-scale 3d domain adaption network for point cloud
  representation.
\newblock In {\em Advances in Neural Information Processing Systems}, pages
  7192--7203, 2019.

\bibitem{Shapira2010}
L.~Shapira, S.~Shalom, A.~Shamir, D.~Cohen-Or, and H.~Zhang.
\newblock Contextual part analogies in 3d objects.
\newblock {\em Int. J. Comput. Vision}, 89(2-3):309--326, Sept. 2010.

\bibitem{sinha2017surfnet}
A.~Sinha, A.~Unmesh, Q.~Huang, and K.~Ramani.
\newblock Surfnet: Generating 3d shape surfaces using deep residual networks.
\newblock In {\em Proceedings of the IEEE conference on computer vision and
  pattern recognition}, pages 6040--6049, 2017.

\bibitem{wang2018pixel2mesh}
N.~Wang, Y.~Zhang, Z.~Li, Y.~Fu, W.~Liu, and Y.-G. Jiang.
\newblock Pixel2mesh: Generating 3d mesh models from single rgb images.
\newblock In {\em ECCV}, 2018.

\bibitem{wang2017cnn}
P.-S. Wang, Y.~Liu, Y.-X. Guo, C.-Y. Sun, and X.~Tong.
\newblock O-cnn: Octree-based convolutional neural networks for 3d shape
  analysis.
\newblock {\em ACM Transactions on Graphics (TOG)}, 36(4):1--11, 2017.

\bibitem{wang20193dn}
W.~Wang, D.~Ceylan, R.~Mech, and U.~Neumann.
\newblock 3dn: 3d deformation network.
\newblock In {\em Proceedings of the IEEE Conference on Computer Vision and
  Pattern Recognition}, pages 1038--1046, 2019.

\bibitem{wang2019voxsegnet}
Z.~Wang and F.~Lu.
\newblock Voxsegnet: Volumetric cnns for semantic part segmentation of 3d
  shapes.
\newblock {\em IEEE transactions on visualization and computer graphics}, 2019.

\bibitem{xiao2020survey}
Y.-P. Xiao, Y.-K. Lai, F.-L. Zhang, C.~Li, and L.~Gao.
\newblock A survey on deep geometry learning: From a representation
  perspective.
\newblock {\em Computational Visual Media}, 6(2):113--133, 2020.

\bibitem{yang2019pointflow}
G.~Yang, X.~Huang, Z.~Hao, M.-Y. Liu, S.~Belongie, and B.~Hariharan.
\newblock Pointflow: 3d point cloud generation with continuous normalizing
  flows.
\newblock In {\em Proceedings of the IEEE/CVF International Conference on
  Computer Vision}, pages 4541--4550, 2019.

\bibitem{yang2018foldingnet}
Y.~Yang, C.~Feng, Y.~Shen, and D.~Tian.
\newblock Foldingnet: Point cloud auto-encoder via deep grid deformation.
\newblock In {\em Proceedings of the IEEE Conference on Computer Vision and
  Pattern Recognition}, pages 206--215, 2018.

\bibitem{yi2016syncspeccnn}
L.~Yi, H.~Su, X.~Guo, and L.~Guibas.
\newblock Syncspeccnn: Synchronized spectral cnn for 3d shape segmentation.
\newblock {\em arXiv preprint arXiv:1612.00606}, 2016.

\bibitem{yin2019logan}
K.~Yin, Z.~Chen, H.~Huang, D.~Cohen-Or, and H.~Zhang.
\newblock Logan: Unpaired shape transform in latent overcomplete space.
\newblock {\em ACM Transactions on Graphics (TOG)}, 38(6):1--13, 2019.

\bibitem{yin2018p2p}
K.~Yin, H.~Huang, D.~Cohen-Or, and H.~Zhang.
\newblock P2p-net: Bidirectional point displacement net for shape transform.
\newblock {\em ACM Transactions on Graphics (TOG)}, 37(4):1--13, 2018.

\bibitem{yumer2016learning}
M.~E. Yumer and N.~J. Mitra.
\newblock Learning semantic deformation flows with 3d convolutional networks.
\newblock In {\em European Conference on Computer Vision}, pages 294--311.
  Springer, 2016.

\bibitem{Zhirong15CVPR}
A.~K. F. Y. L. Z. X. T. J.~X. Z.~Wu, S.~Song.
\newblock 3d shapenets: A deep representation for volumetric shapes.
\newblock In {\em Computer Vision and Pattern Recognition}, 2015.

\bibitem{Zhang2012}
J.~Zhang, J.~Zheng, C.~Wu, and J.~Cai.
\newblock Variational mesh decomposition.
\newblock {\em ACM Trans. Graph.}, 31(3):21:1--21:14, June 2012.

\end{thebibliography}
}

\end{document}